\documentclass[runningheads]{llncs}
\usepackage[T1]{fontenc}
\usepackage{graphicx}
\usepackage{booktabs}
\usepackage[misc]{ifsym}
\newcommand{\corr}{(\Letter)}
\usepackage{xcolor}

\begin{document}

\title{Employing Sentence Space Embedding for Classification of Data Stream from Fake News Domain}

\titlerunning{Employing Sentence Space Embedding for Classification of Data Stream}

\author{
Pawe{\l} Zyblewski\orcidID{0000-0002-4224-6709} \corr \and\\
Jakub Klikowski\orcidID{0000-0002-3825-5514} \and\\
Weronika Borek-Marciniec\orcidID{0000-0003-2426-9541} \and\\
Pawe{\l} Ksieniewicz\orcidID{0000-0001-9578-8395}}
\authorrunning{P. Zyblewski et al.}

\institute{Department of Systems and Computer Networks, 
            Faculty of Information and Communication Technology, 
            Wroc{\l}aw University of Science and Technology,
            Wybrze{\.z}e Wyspia{\'n}skiego 27, 50-370 Wroc{\l}aw, Poland\\ \email{pawel.zyblewski@pwr.edu.pl}}

\maketitle              

\begin{abstract}
Due to the complexity constraints, it is assumed that deep learning methods are not the optimal solution for application in data stream classification tasks. However, excluding the entire -- and prevalent -- group of methods seems rather rash given the progress that has been made in recent years in its development. For this reason, the following paper is the first to present an approach to natural language data stream classification using the sentence space method, which allows for encoding text into the form of a discrete digital signal. This allows the use of convolutional deep networks dedicated to image classification to solve the task of recognizing fake news based on text data. Based on the real-life \emph{Fakeddit} dataset, the proposed approach was compared with \emph{state-of-the-art} algorithms for data stream classification based on generalization ability and time complexity.

\end{abstract}

\section{Introduction}
A widespread opinion among stream learning researchers says that using deep neural networks in this field is a suboptimal solution due to the data processing time~\cite{borisov2022deep}. Excluding the entire pool of solutions characterized often by better quality than canonical methods by assuming too high time complexity as the main argument may be hasty. Thus, the gradual process of deep learning adaptation for data stream processing can be noticed~\cite{duda2020training} in literature, even though currently, it remains an open problem~\cite{borisov2022deep}.

The following work focuses on employing convolutional neural networks to process natural language data streams. The research was based on \emph{state-of-the-art} batch-based methods dedicated to streaming data to account for common problems affecting streams, such as the need for dynamic model updating or the occurrence of concept drift~\cite{komorniczak2023complexity}.

Natural language processing is crucial in the era of general availability of Large Language Models~(\textsc{llm}s). Online portals and social media platforms cause a flood of information and empower the spread of fake news. Analyzing all this data requires stream processing, so models have a chance to adapt to new facts and language dynamics~\cite{ksieniewicz2020fake}. Yet machine learning models cannot process text data in their raw form, which poses an additional challenge to processing time in stream learning.

In conjunction with neural networks for Natural Language Processing (\textsc{nlp}) tasks, embeddings are almost always used~\cite{INCITTI2023418}. It is an advantage of embeddings over canonical methods based on n-grams, which only maintain semantic connection with words in the local context. Nevertheless, in literature, some methods allow for achieving high quality in classifying fake news without using neural networks~\cite{ksieniewicz2023alphabet}.

However, in the end, text preprocessing methods transform it into tabular data. There are models dedicated to text recognition problems that are adapted to such tasks and offer pretrained weights~\cite{devlin2018bert}. However, the pool of such solutions is still smaller than the pool of methods available for image processing, for which convolutional networks are the primary classification tool~\cite{li2021survey}. Methods that transform tabular data into images have already been proposed and used with deep networks with promising results \cite{zyblewski2024employing}. This became the motivation for using this solution for text data in the following paper. For this purpose, the sentence space~\cite{kim2014convolutional} representation was chosen, which allows the creation of an image even in the case of short texts, such as article titles.

Main contributions of this work are: (\textbf{a})~proposal of the \emph{Streaming Sentence Space}~(\textsc{sss}) novelty approach, as an adaptation of sentence space encoding for text data stream classification, (\textbf{b})~development the field of application of deep learning in data stream classification task, which is now considered one of the main research directions related to deep neural networks, and (\textbf{c})~comparison of SSS with \emph{state-of-the-art} data stream ensemble classification algorithms in terms of classification accuracy and computational complexity.

\section{Related works}

This section presents the foundations of the proposed solution, both from the point of view of \textsc{nlp} and data stream processing, and introduces the reference state-of-the-art ensemble classification algorithms dedicated to chunk-based stream processing.

\subsection{Text data extraction methods}

At the core of Natural Language Processing lies the challenge of converting natural language content into a numerical feature space that preserves the document's semantic information. The \emph{bag-of-words} method, a fundamental text feature extraction technique, calculates specific word occurrences within individual samples~\cite{LANG1995331}. This approach forms the basis for more advanced methods such as \emph{bag-of-n-grams}, which retain the contextual information of specific words and thereby a certain semantic association~\cite{furnkranz1998study}. The \textsc{tf-idf} method~\cite{salton1987term} further enhances this vector-based approach by incorporating two key statistics: \emph{Term Frequency}~(\textsc{tf}), which measures word frequency within a document, and \emph{Inverse Document Frequency}~(\textsc{idf}), a logarithmic measure of word uniqueness within a corpus.

In pursuit of the need to reduce dimensionality and the feature vector notation in continuous space, the \emph{continuous bag-of-words} approach was invented, which, along with the \emph{continuous skip-gram} model, is more widely known as \emph{Word2Vec}~(\emph{W2V})~\cite{mikolov2013efficient}. This method allows the determination of a vector representation of a given length for each word in the corpus. Despite its innovation, this approach (\textbf{a})~fails to handle languages that are highly morphologically rich, and (\textbf{b})~determines embeddings only based on local word relationships in the corpus. The answer to the first problem is the \emph{FastText} approach~\cite{10.1162/tacl_a_00051}, which, relying on the \emph{Word2Vec} idea, performs additional word segmentation into \emph{character-n-grams}, significantly enhancing the context during processing. In response to the second drawback, the \emph{Global Vectors} (\emph{GloVe})~\cite{pennington2014glove} model was developed, which is also based on \emph{Word2Vec}, but in addition, extends the model to include general statistics from the processed corpus using a global word-word co-occurrence matrix.

Large Language Models are top-notch among the most commonly used vectorization methods~\cite{INCITTI2023418}. They implement a neural network structure called a transformer as their basis~\cite{NIPS2017_3f5ee243}. Their significant advantage is the employment of self-attention heads that powerfully extend the context, and this, in combination with the processing of massive linguistic resources, generates promising text representations~\cite{10.1145/3605943}. One of the most popular large language models is \emph{Bidirectional Encoder Representations from Transformers}~(\textsc{bert})~\cite{devlin2018bert}, which is designed to pre-train deep bidirectional representations from unlabeled text. As a result, it can capture the dependencies present throughout the text. Additionally, there are no words but subword units called \emph{WordPieces} at the foundation of the \textsc{bert} model. 

However, despite the many advantages of large language models, they have a fair amount of complexity, affecting the time required to determine vectors. A \emph{MiniLM} model~\cite{NEURIPS2020_3f5ee243} is an interesting approach where the authors train a reduced model using knowledge distillation from the \textsc{bert}, maintaining a quality similar to the original.

\subsection{Multi-Dimensional Encoding of text data}
Access to massive volumes of data and increased processing power promotes the usage of deep learning techniques. They are the foundation of multimodal data processing, which is primarily reliant on computer vision tasks, where deep methods frequently outperform canonical approaches~\cite{o2020deep}. Numerous scientific articles confirm that convolutional networks are successfully utilized for image, video, natural language~\cite{gimenez2020semantic}, and audio classification (e.g., in spectrogram form)~\cite{satt2017efficient} in both unimodal and multimodal settings. Deep networks also facilitate transfer learning, enabling models to apply previously learned information to the task at hand~\cite{zhuang2020comprehensive}.

Although the term \emph{Multi-Dimensional Encoding} (\textsc{mde}) is mainly used for tabular data, the \emph{Sentence Space} proposed by Kim~\cite{kim2014convolutional} can be considered as its equivalent for text corpora. Using \emph{sentence space}, text data is transformed into an image in which each row contains embeddings of individual words for each text sample. This approach is clearly dependent on the configuration of the convolutional neural network architecture used, as noted and studied in their work by Zhang and Wallace~\cite{zhang2015sensitivity}, analyzing possible configurations of one-layer \textsc{cnn}s. In turn, Le et al.~\cite{le2018convolutional} analyzed the effect of the depth of convolutional neural networks on \emph{sentence space}. Lately, an extension of the original concept of text encoding to a two-dimensional discrete digital signal was proposed by Soni et al.~\cite{soni2023textconvonet} in the form of \emph{TextConvoNet}. This approach extracts the n-gram features within a sentence and captures the n-gram features between sentences in the input text data, resulting in a three-dimensional representation.

\subsection{Classifier ensemble for imbalanced data stream}
Despite over three decades of research and availability of numerous methods, classifying drifting imbalanced data streams remains one of the important machine learning topics. Methods designed for this task can work in online manner, where each instance is analyzed individually, or on batches of data, where the stream is processed in non-interlacing windows. This work focuses on batch processing, which, because of the larger training set, may provide improved classification quality in the current concept, but is accompanied with a delayed reaction due to waiting for the next data batch~\cite{aguiar2023survey}.

Data imbalance is a prevalent problem in data streams, where the imbalance ratio can be static or dynamic~\cite{aminian2019study}. Most real streams do not have a fixed imbalance ratio, and their properties might change over time~\cite{wang2018systematic}. As a result, data stream classification methods should achieve good classification quality regardless of class distribution, however most techniques built for imbalanced data streams produce unsatisfactory results when class sizes are similar. At the same time, algorithms designed with balanced data in mind, frequently have difficulty with the correct classification of data streams with skewed class distribution~\cite{cano2020kappa}. Methods designed for for dealing with imbalanced data are separated into two main groups: (\textbf{a})~data-level approaches and (\textbf{b})~algorithm-level techniques~\cite{aguiar2023survey}. The first group focuses on data preprocessing, canonically by employing oversampling or undersampling, to change its characteristics prior to classification attempt to alleviate the bias towards majority class, whereas algorithm-level approaches focus on modifying classification algorithms' training phase.

The most prevalent methods for imbalanced data stream classification employ classifier ensembles coupled with data preprocessing techniques. By assuring diversity, constantly updating the classifier pool, and combining the available models, it is feasible to increase the generalization ability and allow for dynamic adaptation to changes in the stream's characteristics~\cite{brzezinski2018ensemble}. Among the established algorithms, we can distinguish \emph{Concept Drift with Smote}~(\emph{Learn++.CDS}) and \emph{Nonstationary and Imbalanced Environments}~(\emph{Learn++.NIE}) by Ditzler and Polikar~\cite{Ditzler:2013}. \emph{Learn++.CDS} extends the \emph{Learn++.NIE} algorithm by employing the \textsc{smote} in attempt to balance the number of samples in each class, while \emph{Learn++.NIE} utilizes a penalty constraint to balance classification accuracy on all classes, while also employing a bagging-based sub-ensemble. Wang et al.~\cite{Wang2015} proposed the \emph{Oversampling Online Bagging} (\textsc{oob}) and \emph{Undersampling Online Bagging} (\textsc{uob}) algorithms dedicated for online data stream processing, extending Online Bagging by altering the Poisson distribution $\lambda$ parameter according to the current imbalance ratio. Cano and Krawczyk~\cite{cano2020kappa} developed \emph{Kappa Updated Ensemble} (\textsc{kue}), which combines batch-based and online processing on feature subspaces. \textsc{kue} uses the Kappa statistic to dynamically weight and select base classifiers. The same authors introduced also the \emph{Robust Online Self-Adjusting Ensemble}~(\textsc{rose}) for non-stationary data stream classification~\cite{cano2022rose}. This method trains online learners based on data views, ensuring a diverse classifier pool. It also employs drift detectors to respond quickly to changes in data distribution and proposes effective strategies for dealing with data imbalance. Woźniak et al. used built-in mechanisms (e.g., weighting and aging of classification models) to establish a self-updating classifier pool that can adjust its lineup in response to changes in imbalance ratio and concept drift \cite{wozniak2023active}. Klikowski and Woźniak~\cite{klikowski2020employing} trained one-class classifiers using clustered data, while Zyblewski et al.~\cite{zyblewski2021preprocessed} proposed to combine Dynamic Classifier Selection with data preprocessing techniques for imbalanced data stream classification.

\section{Streaming Sentence Space}

\begin{figure}[!htb]
  \begin{center}
    \includegraphics[width=0.99\columnwidth,trim={1.1cm 23.5cm 7.2cm 1.1cm},clip]{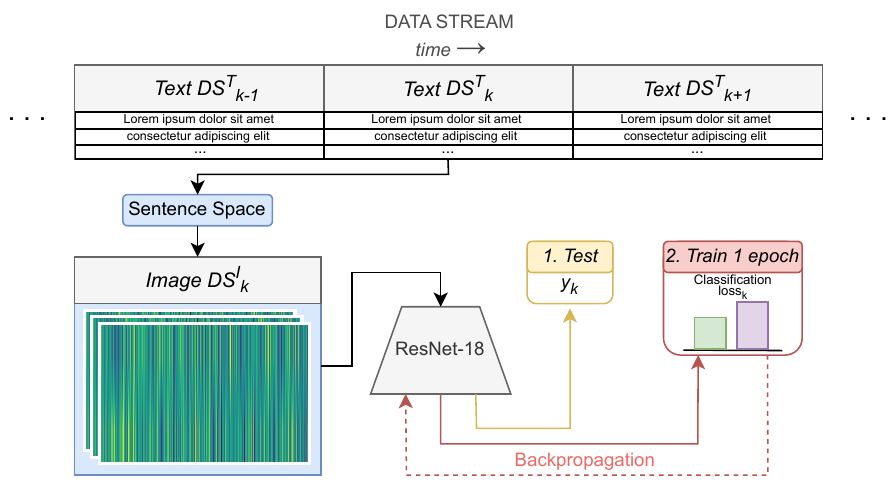}
  \end{center}
  \caption{The general scheme of the proposed SSS approach.}
  \label{fig:sss}
\end{figure}

As observed in Section 1, solutions based on deep learning are often overlooked in data stream classification tasks. This is due to concerns about increased computational and time complexity, in both the induction and inference process~\cite{michalski1993inferential}, despite the tremendous recent progress made in the area of deep learning. One promising solution is \emph{Sentence Space}, a \emph{multi-dimensional encoding} counterpart for text data. While attempts have already been made in the literature to use \textsc{mde} approaches for tabular data in data stream scenarios \cite{zyblewski2024employing}, despite numerous works confirming the performance of \emph{sentence space} and its derivatives, no studies analyzing the use of this encoding in the task of classifying streams containing text data have been produced so far. In order to fill this niche, this paper proposes \emph{Streaming Sentence Space}~(\textsc{sss}), thus taking the first step toward analyzing the application of sentence space in the classification of dynamically imbalanced data streams from fake news domain.

The main assumption behind this work was to keep the time complexity as low as possible and to enable the use of \textsc{sss} in real-life data stream classification tasks while maintaining the generalization potential inherent to convolutional neural networks. Accordingly, this work analyzes only a batch-based processing scenario, in which prediction and model training are performed on a window of predefined size. The need to wait for a single data chunk to fill up, depending on the dynamics of the data stream, can significantly reduce problems arising from possibly increased processing time.

The basis of \textsc{sss} is the conventional \emph{sentence space} encoding~\cite{kim2014convolutional}, in which individual words are transformed into embeddings that represent consecutive lines of an image. Of course, it is also possible to use approaches such as \emph{TextConvoNet}, but this depends on the characteristics of the data being analyzed. The decision to use sentence space in this case was related to the relative short length of the texts contained in the stream corpus (more in Section~4). As for the convolutional network, the decision was made to use the popular \emph{ResNet-18}~\cite{he2015deepresiduallearningimage} architecture with the assumption of only one training epoch in each successive data chunks. The standard and commonly employed optimizer \textsc{sgd} with learning rate of 0.001 and momentum of 0.9 was used, batch size was set to 8. \emph{Cross-entropy loss} was used as the loss function.

Figure~\ref{fig:sss} illustrates \textsc{sss}-based data stream processing. We regard the data stream as a sequence of text data chunks $DS^T_k$ with a fixed size of $N$, where $k$ is the batch index. \textsc{sss} encodes each incoming data chunk into a series of two-dimensional discrete digital signals $DS^I_k$ that contain $N$ pictures with a predetermined side size. Each $N$ picture from $DS^I_k$ is copied three times to provide an image representation with three color channels for the \emph{ResNet-18} architecture. \emph{ResNet-18} follows the \emph{Test-Then-Train protocol}, performing inference and one training epoch for each data batch $DS^I_k$.

\begin{figure}[!htb]
    \centering
    \includegraphics[width=.99\textwidth]{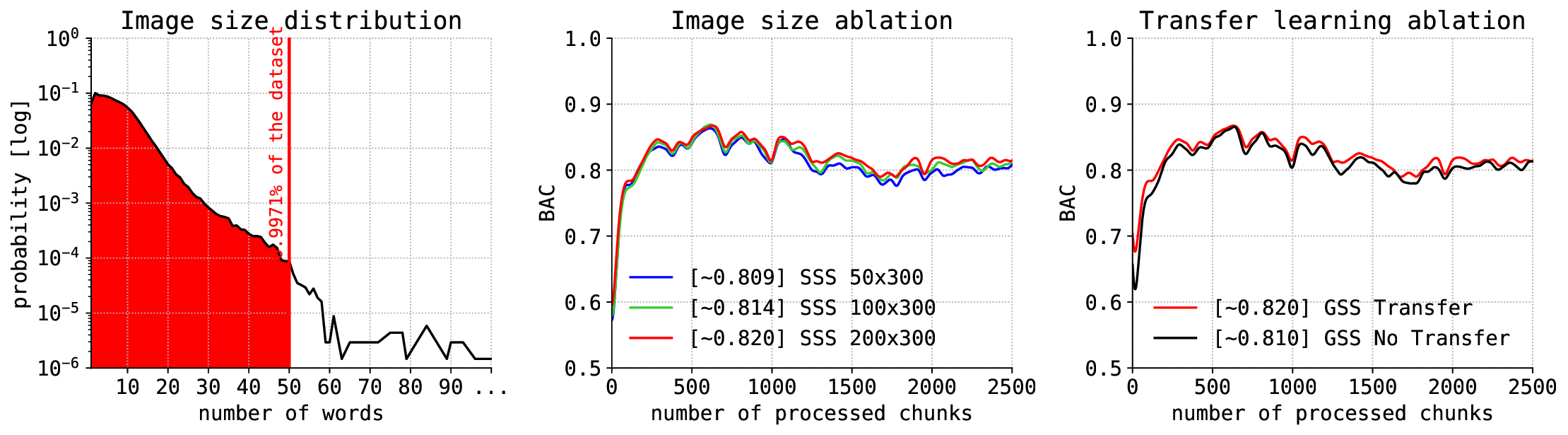}
    \caption{Results of preliminary experiments related to image size and transfer learning.}
    \label{fig:0_preliminary}
\end{figure}

One of the fundamental problems arising from \emph{sentence space} is the approach to determining the dimensions of the images resulting from encoding. The width is typically the length of the embedding vector, but selecting the height is not a trivial task, as it depends on the number of words contained in the text. Approaches based on padding or clipping texts to a specific length can be used, but in the case of \textsc{sss}, it was decided to resize the heights of the images by bilinear interpolation. The first two subplots of Figure~\ref{fig:0_preliminary} show the experimental process of selecting image heights for the \emph{fakeddit} dataset. Text embeddings were obtained using the \emph{GloVe} technique, which is a more recent alternative to \emph{Word2Vec} usually used for this purpose and offers better recognition quality in many problems. After analyzing the distribution of the number of words in the corpus texts, it was found that almost all of them were in the range of up to 50 words in length. Due to this observation, 50x300 px was set as the initial dimensions of sentence space images after resize. In addition, this experiment was repeated for dimensions of 100x300 px and 200x300 px. The obtained values of \emph{balanced accuracy score}, although very close, indicate the advantage of images with a height of 200 px, and therefore this is the value used in the experiments presented next.

In addition, due to the relatively unusual characteristics of the resulting images of sentence space encoding and the possibility of negative knowledge transfer, a short experiment was conducted to determine the validity of using the \emph{ResNet-18} architecture pretrained on the \emph{ImageNet} dataset. In the image height experiment, transfer learning was applied by default, as is the case in many research articles, but in this case it was decided to repeat the study using the \emph{ResNet-18} model architecture learned from scratch. The results obtained, presented in the last subplot of Fig.~\ref{fig:0_preliminary}, indicate a minimal advantage of the pretrained network when classifying images resulting from sentence space encoding.

\section{Experimental Evaluation}
The experimental study conducted to evaluate the performance of the \textsc{sss} was designed to answer the following research questions: 
\begin{itemize}
    \item \textbf{RQ1} Which of the commonly used approaches for obtaining representations from text data for pattern recognition tasks should be used in conjunction with \emph{sentence space} to obtain images that allow \textsc{cnn}s to achieve the highest generalization capability?
    \item \textbf{RQ2} Does the use of \textsc{sss} make it possible to achieve classification quality superior to \emph{state-of-the-art} ensemble data stream classification algorithms trained using representations obtained from commonly used extractors?
    \item \textbf{RQ3} How does the time complexity of \textsc{sss} compare to \emph{state-of-the-art} ensemble data stream classification algorithms, and does it enable its use in real-life data stream classification tasks?
\end{itemize}

\subsection{Set-up}

\begin{figure}[!htb]
    \centering
    \includegraphics[width=.88\textwidth]{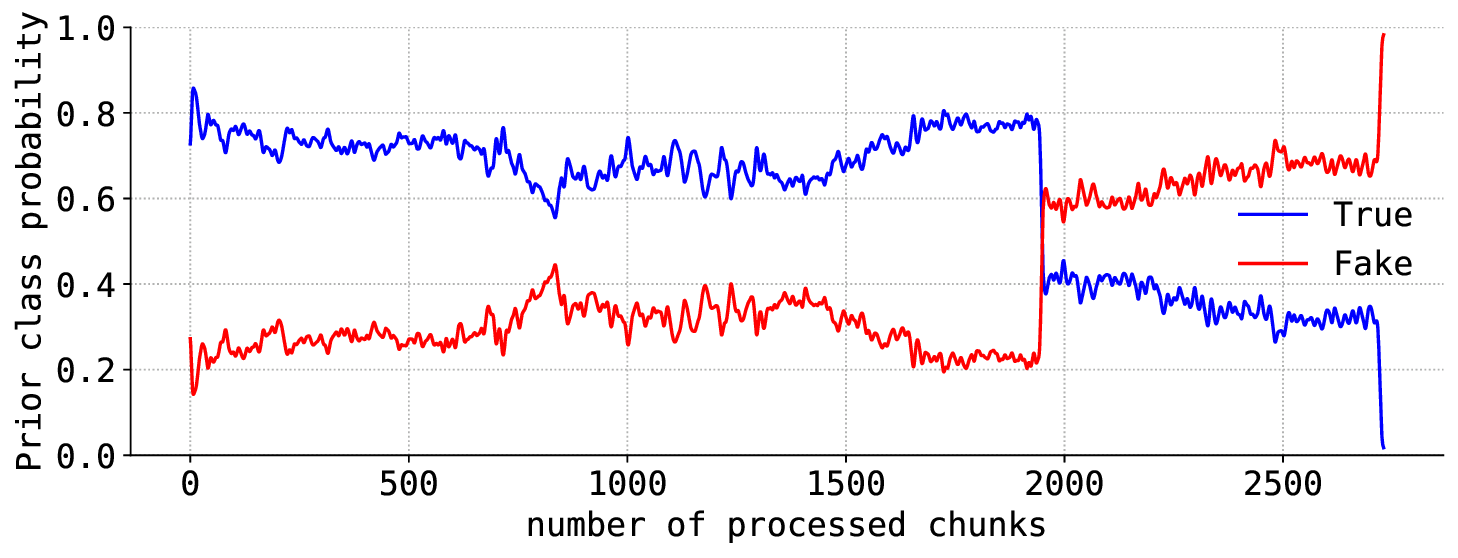}
    \caption{Changes in the prior class probabilities over time.}
    \label{fig:prior}
\end{figure}

\textbf{Data} All of the research was conducted using \emph{Fakeddit's multimodal dataset}, which presents a real-life fake news classification task broken down into two, three or six classes~\cite{nakamura2019r}. This dataset consists of more than one million posts on 22 different subreddits of the social networking platform \emph{Reddit} and includes text and image modalities, supplemented by metadata about the posts and their authors. For the purposes of this study, a single binary data stream was prepared, in which consecutive texts were sorted accordingly to their creation timestamp. Of the entire dataset, $682,996$ multimodal samples were used (the rest have only one modality). The decision to limit to multimodal samples only is linked to the facilitation of extending the presented research to include the image modality. The data stream was divided into $2731$ data chunks, containing $250$ samples each. The resulting data stream is characterized by a dynamic imbalance ratio, which changes in successive batches to the point where, at about 3/4 of the stream length, a minority class transitions into a majority class. The changes in the prior probability of class membership are shown in Fig.~\ref{fig:prior}.

\textbf{Experimental protocol \& reproducibility} All experiments were carried out using the \emph{Test-Then-Train} protocol to guarantee a robust experimental evaluation, were implemented in \emph{Python} and can be replicated using the publicly available \emph{GitHub} repository\footnote{\url{https://github.com/w4k2/sentence-space-stream}}. Implementation of \emph{state-of-the-art} algorithms were based on \emph{stream-learn}~\cite{ksieniewicz2022stream}, \emph{scikit-multiflow}~\cite{montiel2018scikit}, and \emph{PyTorch}~\cite{paszke2017automatic} libraries. The classification quality evaluation of the algorithms was based on the standard metrics used in the task of imbalanced data classification, i.e. \emph{balanced accuracy score}~(\textsc{bac}), \emph{recall}, \emph{specificity}, \emph{precision}, $F_1$ score, $Gmean$, and $Gmean_s$.

\subsection{Experiment scenarios}

\noindent\textbf{Experiment 1 -- Extraction methods} The goal of Experiment~1 was to compare the performance of \textsc{sss} depending on the type of extractor used to obtain a representation for Sentence Space encoding. For this purpose, (\textbf{a})~\emph{GloVe}, (\textbf{b})~\emph{MiniLM}, (\textbf{c})~pretrained \emph{Word2Vec}, and (\textbf{d})~\emph{Word2Vec} updated after each data chunk were compared with each other. The representation width depending on the extractor was $380$ for \emph{MiniLM} and $300$ for \emph{GloVe} and \emph{W2V}. Based on the results, the feature extraction method used in subsequent experiments was selected.

\noindent\textbf{Experiment 2 -- Comparison with data stream classification algorithms} In Experiment~2, the \textsc{sss} based on the extractor chosen in Experiment~1 was compared with \emph{state-of-the-art} ensemble algorithms for imbalanced data stream classification. Among the methods mentioned in the literature review, (\textbf{a})~\emph{Hoeffding Tree} with Hellinger split criterion~(\textsc{hf}), (\textbf{b})~\emph{Learn++.CDS}~(\textsc{cds}), (\textbf{c})~\emph{Learn++ .NIE}~(\textsc{nie}), (\textbf{d})~\emph{Kappa Updated Ensemble}~(\textsc{kue}), and (\textbf{e})~\emph{Robust Online Self-Adjusting Ensemble}~(\textsc{rose}) were selected as references. \textsc{hf} was used as the base classifier for all reference methods, and the maximum size of the classifier pool was set to $10$. The selection of methods, base classifier, and pool size was based on the literature~\cite{aguiar2023survey,cano2020kappa,cano2022rose}. The entire set of reference algorithms was compared with the \textsc{sss} depending on the approach used to extract features from the text. In addition to the extractors used in Experiment~1, \textsc{tf-idf} with unigrams and bigrams and $100$ features with top term frequency was employed here. A set of reference methods trained using the representation that provided the highest classification quality in terms of \textsc{bac} was selected for the last experiment.

\noindent\textbf{Experiment 3 -- Time complexity}
The last experiment was designed to analyze the emerged methods in terms of time complexity. For this purpose, for both \textsc{sss} and reference methods, the feature extraction, prediction and training times for the first $110$ data chunks from the \emph{fakeddit} stream were measured, respectively. To account for the processing time of reference methods only for the classifier pool with the maximum number of models, the first $10$ data chunks were ignored. The experiment was repeated $10$ times to stabilize the results obtained.

\subsection{Experiment 1 -- Extraction methods}
\begin{figure}[!htb]
    \centering
    \includegraphics[width=.8\columnwidth]{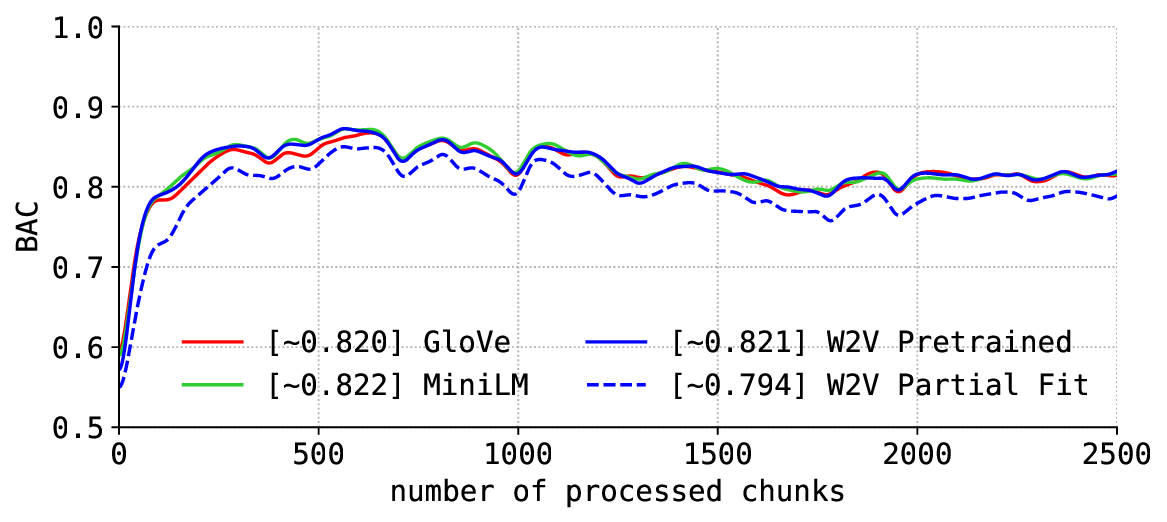}
    \caption{Results of an experiment to determine the best extraction method for SSS.}
    \label{fig:1_extractor}
\end{figure}

As can be seen in Figure~\ref{fig:1_extractor}, across all data chunks, the results of the first three methods -- \emph{GloVe}, \emph{MiniLM}, and \emph{Word2Vec pretrained} -- are similar, with the only deviating method being \emph{Word2Vec partial-fit} trained on each chunk. This discrepancy can be explained by the dictionary's limitations resulting from training on successive data chunks containing only short texts. In contrast, \emph{MiniLM}, \emph{Word2Vec pretrained} (\emph{word2vec-google-news-300}), and \emph{GloVe}~(\emph{glove-wiki-gigaword-300}), were all built on larger volumes of data compared to the \emph{Word2Vec} trained from scratch.

Following this experiment, it was decided to use the \emph{GloVe} vectors for further research, which is a newer method than \emph{Word2Vec} that processes additional global information and, at the same time -- unlike \emph{MiniLM}, dedicated to sequential processing -- was designed to process single words. In addition, the \emph{MiniLM} model requires more computational complexity when determining the representation vector for a given word, which for \emph{Word2Vec} and \emph{Glove} methods is limited only to reading values from the vector array. With equal effectiveness, the exclusion of \emph{MiniLM} is fully justified. Thus, the results obtained in Experiment 1 made it possible to answer \textbf{RQ1}.

\subsection{Experiment 2 -- Comparison with data stream classification algorithms}
The preliminary part of the main comparative experiment begins with analyzing the influence of feature space reduction, which was achieved using \emph{Principal Component Analysis}~(\textsc{pca}) -- projecting the original embedding space down into $100$ features. 
Based on the achieved results, it can be concluded that the influence of this simple reduction on the results is mostly cosmetic, always laying in a one percent margin of a difference, so it is justified to reduce problem representation for canonical models since representation gets smaller while change in classification quality is negligible.=

\begin{figure}[!htb]
    \centering
    \includegraphics[width=.99\textwidth]{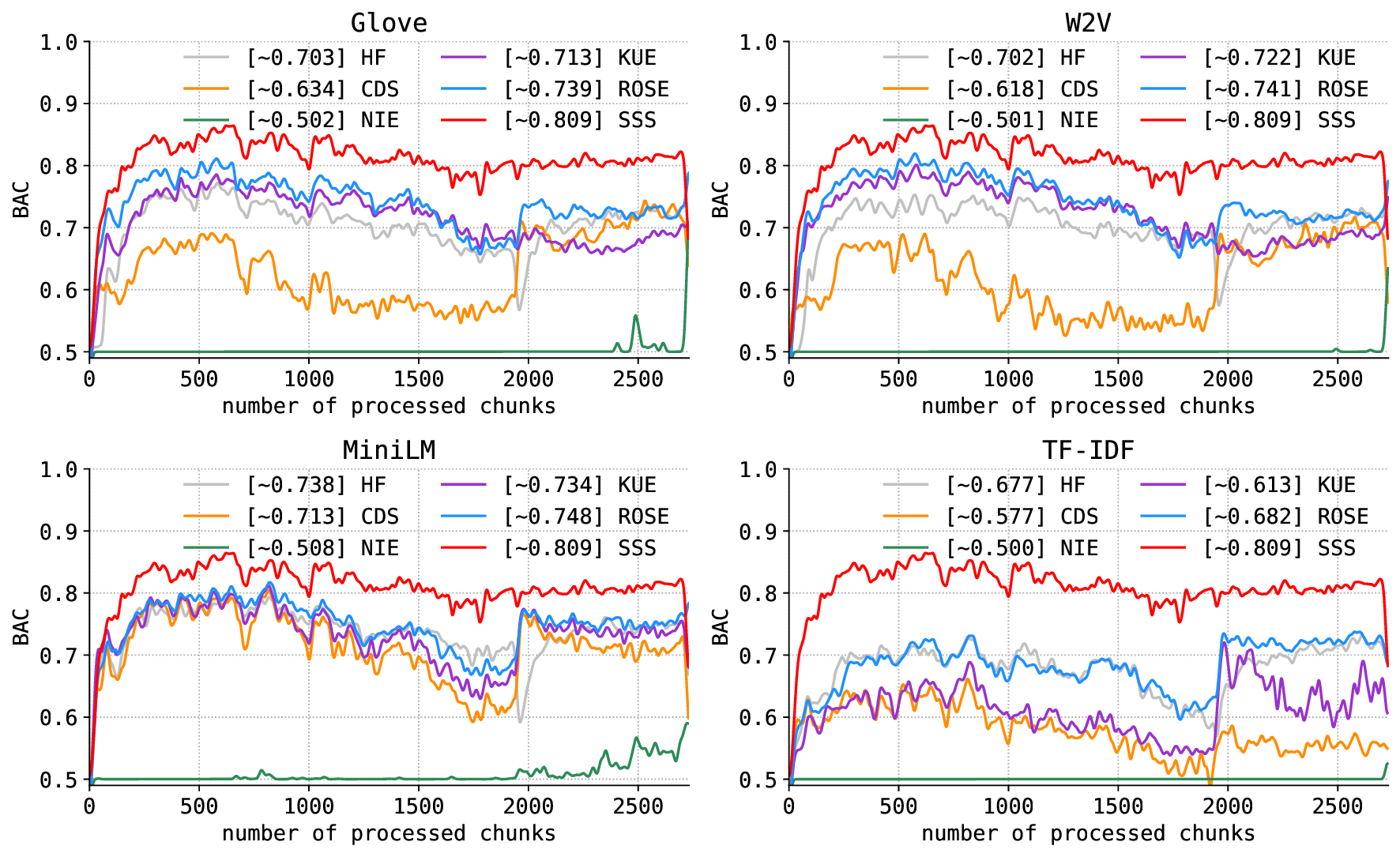}
    \caption{Comparison of SSS with reference methods depending on the extraction method used.}
    \label{fig:2_comparison}
\end{figure}

The experimental evaluation results, taking into account the proposed \textsc{sss} approach, are presented in Figure~\ref{fig:2_comparison}, divided into four subfigures according to the extraction strategies used for canonical methods. Regardless of the extraction method, the \textsc{nie} algorithm performs the worst, maintaining \textsc{bac} at the level of a random classifier for almost the whole stream length. Only when using \emph{Glove} and \emph{MiniLM} embeddings does its balanced accuracy score rise slightly in the final phase of the stream, after the prior concept drift shown in Fig.~\ref{fig:prior}.

The second weakest method is \emph{Learning++CDS}, which, for all extractors except \emph{MiniLM}, clearly leans towards randomness. All analyzed reference solutions appear to be very sensitive to concept drift occurring around the chunk $2,000$. The ranking of \textsc{kue}, \textsc{rose}, and \emph{Hoeffding Tree} algorithms depends on the extraction strategy used. However, their effectiveness's overall distribution is similar, with a slight advantage for \textsc{rose}.

\begin{figure}[!htb]
    \centering
    \includegraphics[width=.65\textwidth]{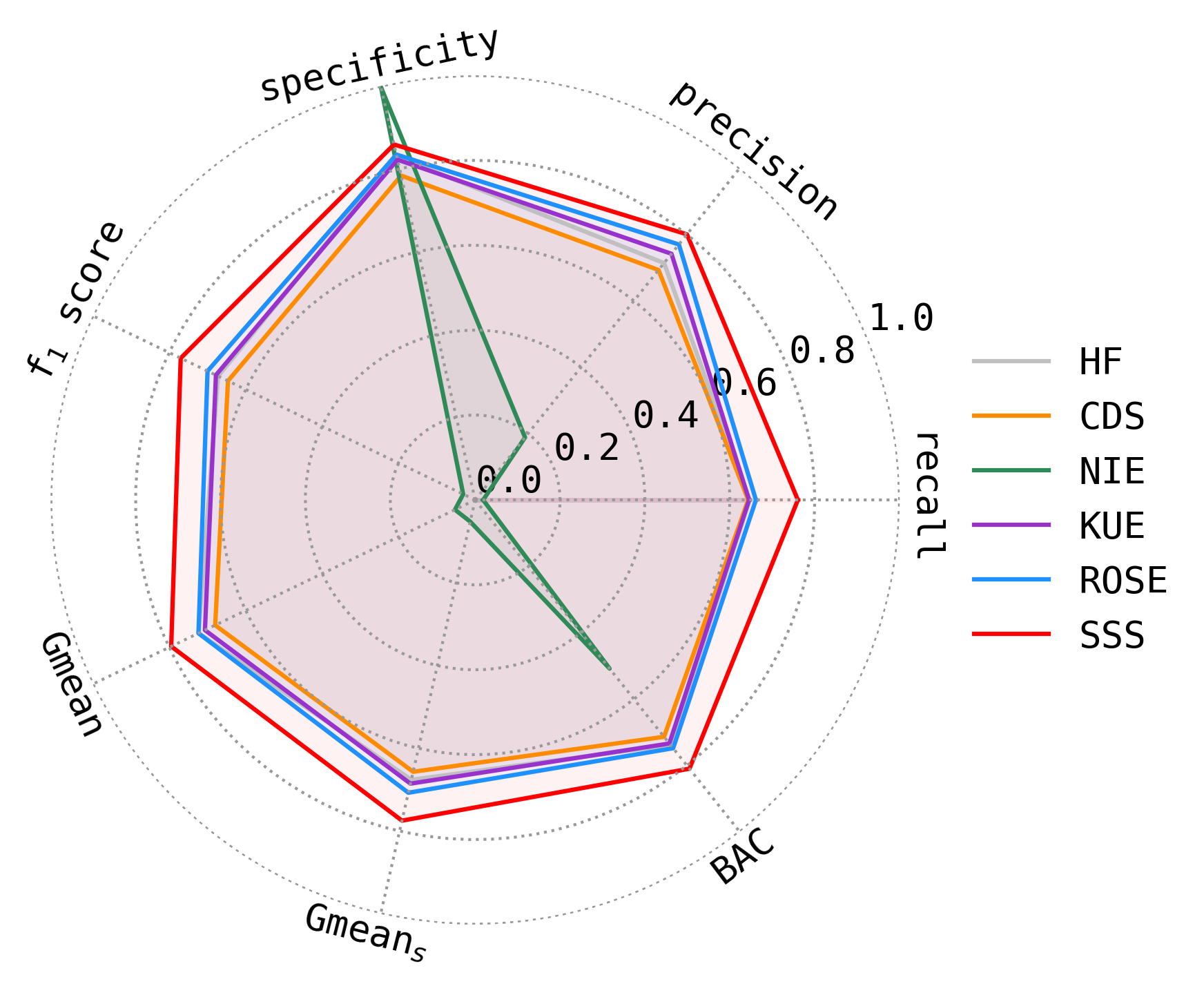}
    \caption{Comparison of SSS with reference methods trained using MiniLM embeddings.}
    \label{fig:4_radar}
\end{figure}

\textsc{sss} shows a noticeable advantage over all canonical solutions throughout the entire data stream, being the only one to maintain an average recognition efficiency of $80\%$ \emph{balanced accuracy score} (thereby responding to \textbf{RQ2}). This observation is supported by the extended analysis of metrics in the form of a radar diagram (Figure~\ref{fig:4_radar}), where an advantage of \textsc{sss} can be observed in each of the base and aggregated metrics used. The outlier \emph{specificity} result for the \textsc{nie} method stems from its complete inability to learn in the analyzed problem environment, which for the dominant majority of processing time induces decisions towards only one of the problem classes, without a generalized relation to given bias. 

Among the analyzed extraction methods, \emph{MiniLM} comes minimaly to the fore, as it was the only one that allowed the \textsc{nie} strategy to noticeably rise from the random classifier level in the final part of the stream, and the \textsc{cds} to compete with the rest of the algorithms. Accordingly, it was the \emph{MiniLM} embeddings that were used for the canonical methods in Experiment 3.

\subsection{Experiment 3 -- Time complexity}

\begin{figure}[!htb]
    \centering
    \includegraphics[width=.99\textwidth]{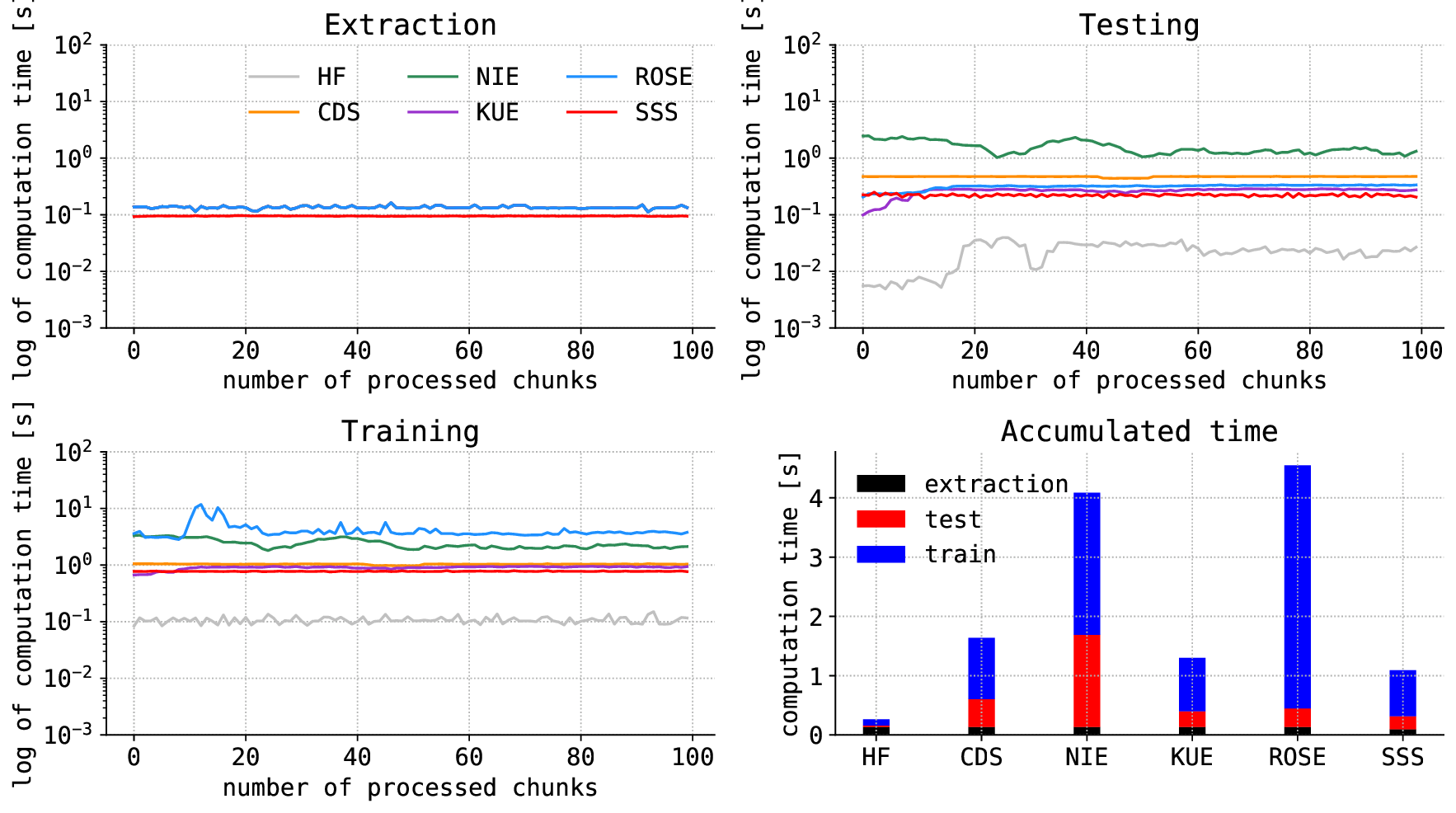}
    \caption{Comparison of SSS with reference methods in terms of time complexity.}
    \label{fig:5_time}
\end{figure}

The results of the third experiment -- showing the time complexity for extracting, training, and testing the algorithm -- are shown in Figure~\ref{fig:5_time}. As we can see, in the case of preprocessing, all reference methods show uniform computation time (lines overlap), and only the proposed approach deviates from this tendency and performs extraction faster. Additionally, it should be noted that this happens despite the time measurement considering the transition to image representation. The method owes it using the \emph{GloVe} technique for \textsc{sss}, which proved to work best in Experiment~1. The reference methods, however, use the \emph{MiniLM} transformer in accordance with the outcome of Experiment~2.

More variability can be observed in the training and testing processes -- in both, a single \emph{Hoeffding Tree} processes the fastest. In turn, training in a single epoch is the slowest for \textsc{rose}, and prediction -- for \textsc{nie}. It is also reflected in the accumulated time graph, which indicates that the only method ahead of \textsc{sss} regarding the entire processing time is the \emph{Hoeffding Tree}. Still, it should be noted that a single classifier will always be faster than an ensemble.

Therefore, the proposed method is faster then the classifier ensembles used for data streams even though it is based on convolutional network and requires encoding from text into an image. At the same time, despite the lowest processing time compared to \emph{state-of-the-art} ensemble algorithms, \textsc{sss} offers the highest generalization ability. All this means that \textsc{sss} can be successfully used in real-life batch-based data stream classification tasks, answering the \textbf{RQ3}.

\section{Conclusions}
The presented research work aimed to achieve two main goals. The first was to address the application of deep learning in the task of data stream classification, which is presented in the current literature as one of the main research areas in the need of further investigation. The second goal was to propose the use of \emph{sentence space}, which is the equivalent of \emph{multi-dimensional encoding} for text data, in a data stream classification task.

To realize the above goals, \emph{Streaming Sentence Space}~(\textsc{sss}) was proposed, which encodes the text found in individual data batches into discrete digital signals based on embeddings obtained through the \emph{GloVe} technique. The resulting images are then classified using the \emph{ResNet-18} architecture, which, in order to reduce computational complexity, performs only a single training epoch on each data chunk. The developed approach was tested on the basis of computer experiments conducted on a real-life dynamically imbalanced data stream formed by chronologically ordering the texts contained in the \emph{fakeddit} dataset. The results showed that \textsc{sss}, thanks to the inherent generalization ability of the convolutional neural network, is able to outperform the classification quality of \emph{state-of-the-art} classifier ensemble methods dedicated for imbalanced data stream classification. In addition, \textsc{sss} exhibits lower time complexity than ensemble reference methods, which further encourages its use and contradicts the popular opinion that deep learning has too high time and computational complexity to be used for data stream analysis.

Future research may focus on examining the applicability of other sentence space derived techniques for encoding text into image form in the task of data stream classification. Another potentially interesting direction is the application of \emph{sentence space}-based approaches in the classification of multimodal data streams containing text modality.

\begin{credits}
\subsubsection{\ackname} 
This work was supported by the National Center for Research and Development within INFOSTRATEG program, number of application for funding: INFOSTRATEG-I/0019/2021-00.

\end{credits}
%
\bibliographystyle{splncs04}
\bibliography{bibliography}

\end{document}